\documentclass[10pt,twocolumn,letterpaper]{article}

\usepackage[pagenumbers]{cvpr}      

\usepackage{xcolor,colortbl,soul}
\usepackage{seqsplit}
\usepackage[scaled=0.85]{inconsolata}

\newcommand{\IGNORE}[1]{}

\definecolor{vermilion}{rgb}{0.89, 0.26, 0.2}

\definecolor{cvprblue}{rgb}{0.21,0.49,0.74}
\usepackage[pagebackref,breaklinks,colorlinks,allcolors=cvprblue]{hyperref}
\usepackage{xcolor}

\def\confName{CVPR}
\def\confYear{2026}

\title{CamLit: Unified Video Diffusion with Explicit Camera and Lighting Control}

\author{
  Zhiyi Kuang$^{1,2}$ \quad
  Chengan He$^{1}$ \quad
  Egor Zakharov$^{1}$ \quad
  Yuxuan Xue$^{1,3}$ \quad
  Shunsuke Saito$^{1}$ \\
  Olivier Maury$^{1}$ \quad
  Timur Bagautdinov$^{1}$ \quad
  Youyi Zheng$^{2}$ \quad
  Giljoo Nam$^{1}$ \\
  $^{1}$Codec Avatars Lab, Reality Labs, Meta \\
  $^{2}$State Key Lab of CAD\&CG, Zhejiang University \\
  $^{3}$University of T{\"u}bingen
}

\begin{document}


\twocolumn[{
    \renewcommand\twocolumn[1][]{#1}%
    \maketitle
    \begin{center}
        \includegraphics[width=\textwidth]{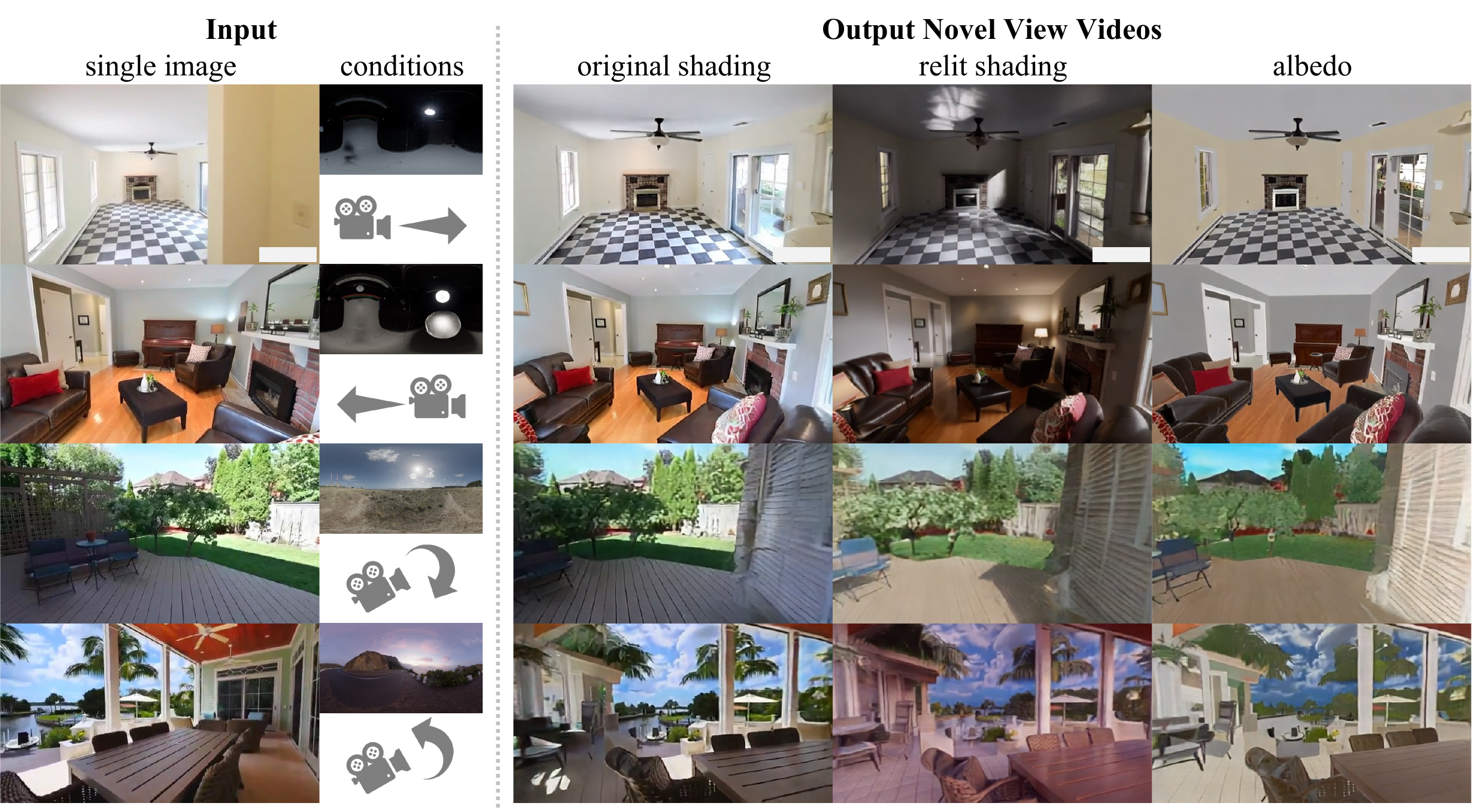}
        \captionof{figure}{\textbf{CamLit, a unified video diffusion model with joint camera and lighting control.} Given a single image, CamLit generates a novel view video, a paired relit video, and a paired albedo video under user-defined camera trajectory and lighting conditions with high fidelity.}
        \label{fig:teaser}
    \end{center}
}]

\begin{abstract}
We present CamLit, the first unified video diffusion model that jointly performs novel view synthesis (NVS) and relighting from a single input image. Given one reference image, a user-defined camera trajectory, and an environment map, CamLit synthesizes a video of the scene from new viewpoints under the specified illumination. Within a single generative process, our model produces temporally coherent and spatially aligned outputs, including relit novel-view frames and corresponding albedo frames, enabling high-quality control of both camera pose and lighting. Qualitative and quantitative experiments demonstrate that CamLit achieves high-fidelity outputs on par with state-of-the-art methods in both novel view synthesis and relighting, without sacrificing visual quality in either task. We show that a single generative model can effectively integrate camera and lighting control, simplifying the video generation pipeline while maintaining competitive performance and consistent realism.
\end{abstract}    
\section{Introduction}
\label{sec:intro}
%
Modern vision systems, from robot perception to augmented reality, require large volumes of video data that capture 3D camera motion under diverse illumination. To be robust, they must detect objects and recover 3D geometry even in challenging lighting, such as strong specular reflections or cast shadows. However, collecting videos with diverse camera motions and lighting conditions is costly. In contrast, single images are abundant. This motivates us to build a data augmentation tool that converts a single static image into photorealistic video sequences with explicit control over camera trajectory and lighting conditions, providing a scalable way to train models that are resilient to changes in viewpoint and illumination.

We introduce \emph{CamLit}, a unified video diffusion model that enables explicit control over both camera motion and lighting. To the best of our knowledge, CamLit is the first framework to jointly perform novel view synthesis (NVS) and relighting within a single model. Given a single input image, a user-defined camera trajectory, and an environment map, CamLit generates a video of the scene as if it were captured along the specified trajectory and under the designated illumination. This capability unlocks vast potential for data augmentation in training and simulation: for example, a single indoor snapshot can produce a realistic room tour rendered under diverse lighting conditions.

At the core of CamLit is a multi-modal video Diffusion Transformer (DiT)~\cite{peebles2022dit}. This model takes as input a single RGB image (defining the scene content), a camera trajectory (defining the viewpoint at each video frame), and an HDR environment map (defining the incident illumination). From these inputs, CamLit generates a spatially and temporally aligned triplet of videos: (i) an RGB novel-view sequence under the same illumination as the input image, (ii) the corresponding relit sequence (with full shading from the environment map), and (iii) an albedo sequence capturing the scene’s intrinsic colors without shading. As demonstrated in~\cite{mei2025lux,he2025unirelight}, this design enforces cross-modal coherence: the model learns a common implicit scene representation that ensures shading effects such as cast shadows in the input image to be effectively removed in the relit video.

However, a key challenge for this joint denoising formulation is the lack of paired multi-view, multi-illumination video triplets for training. To address this, we curate a large-scale dataset of paired videos. We leverage the RealEstate10K dataset~\cite{zhou2018stereo} of in-the-wild videos and an existing neural renderer to generate training supervision. Specifically, we process $56,975$ real video clips from RealEstate10K with DiffusionRenderer~\cite{DiffusionRenderer}, yielding a large number of triplets of (original video, albedo video, relit video) for diverse scenes, which we use to train our diffusion model. 

In summary, CamLit elevates single-image content into controllable videos, unifying camera and lighting control in a single diffusion-based generative model. Experiments show that CamLit achieves high-quality generation on indoor and outdoor scenes, producing plausible novel views and relighting effects without sacrificing fidelity in either task.
\section{Related Work}
\label{sec:related_works}

\paragraph{Novel View Synthesis from Sparse Inputs.}
A straightforward approach to novel view synthesis (NVS) from single or sparse images involves first reconstructing the 3D scene geometry, then rendering novel viewpoints from this representation~\cite{chen2024mvsplat,yu2021pixelnerf,xu2025depthsplat,ziwen2025long,charatan2024pixelsplat,liu2023zero,sargent2024zeronvs,wu2023reconfusion,voleti2024sv3d,yu2024viewcrafter}. 
However, recent work has demonstrated that bypassing intermediate 3D representations can yield more scalable and generalizable solutions. Large View Synthesis Model (LVSM)~\cite{jin2024lvsm} exemplifies this approach through a transformer-based feed-forward network that directly generates novel-view images from input images and target camera parameters, eliminating the need of 3D inductive bias during novel view synthesis.
This direct generation strategy has been further advanced by diffusion-based NVS methods~\cite{he2024cameractrl,watson2024controlling,wang2024motionctrl,zhou2025stable}, which condition the denoising process of image and video diffusion models on target camera parameters.
While these diffusion-based approaches achieve impressive results for single-image NVS, they lack explicit control over scene appearance under different lighting conditions, which is the critical limitation addressed in this paper.

\paragraph{Image and Video Relighting.}
When talking about relighting, a classical perspective is to rely on reconstructing scene geometry and reflectance parameters (\eg, SVBRDF), followed by re-rendering using the physically-based rendering equation~\cite{kajiya1986rendering}. This inverse rendering pipeline has been the dominant paradigm for image relighting in computer graphics~\cite{boss2021nerd,physg2021,rudnev2022nerfosr,DIBR19,zhang2021nerfactor}. Yet with the emergence of diffusion models~\cite{ho2020denoising}, recent advances have shifted toward diffusion-based generative models for image and video relighting~\cite{zhang2025scaling,zeng2024dilightnet,bharadwaj2024genlit,kocsis2024lightit,zeng2024rgb,diffusion_relighting,xing2024luminet}. These methods bypass explicit material appearance modeling and physically-based rendering, which often constrain photorealism for objects with complex optical properties.
A key breakthrough along this direction has been the integration of environment maps into the diffusion process~\cite{chaturvedi2025synthlight,jin2024neural_gaffer,DiffusionRenderer,he2025unirelight}, enabling precise control over lighting conditions. Most notably, UniRelight~\cite{he2025unirelight} demonstrated that joint denoising of albedo and relit frames allows the model to learn realistic lighting effects including shadows and reflections. While our framework draws inspiration from UniRelight's joint denoising strategy, we extend this approach beyond relighting to simultaneously perform novel view synthesis, enabling unified control over both viewpoint and illumination.

\paragraph{Multimodal Diffusion Models.}
Recent advances demonstrate that jointly denoising multiple modalities within a single diffusion process significantly improves cross-modal coherence and generalization compared to independent generation approaches. This paradigm has proven effective across diverse domains, including audio-visual generation~\cite{ruan2022mmdiffusion}, vision-language modeling~\cite{li2025dual}, 3D reconstruction~\cite{lu2025matrix3d}, video generation~\cite{hila2025videojam}, and video relighting~\cite{he2025unirelight}. The typical architectural strategy involves concatenating heterogeneous modality tokens or latent codes into a unified transformer model, enabling full cross-modal attention to capture inter-modal dependencies.
Our framework adopts this joint denoising paradigm by simultaneously generating three complementary outputs: (i) novel-view frames that preserve the original scene appearance, (ii) relit frames rendered under user-specified environment lighting, and (iii) corresponding albedo maps that capture intrinsic material properties. 
\section{Methodology}
\label{sec:method}

\begin{figure*}[htbp]
    \centering
    \includegraphics[width=\textwidth]{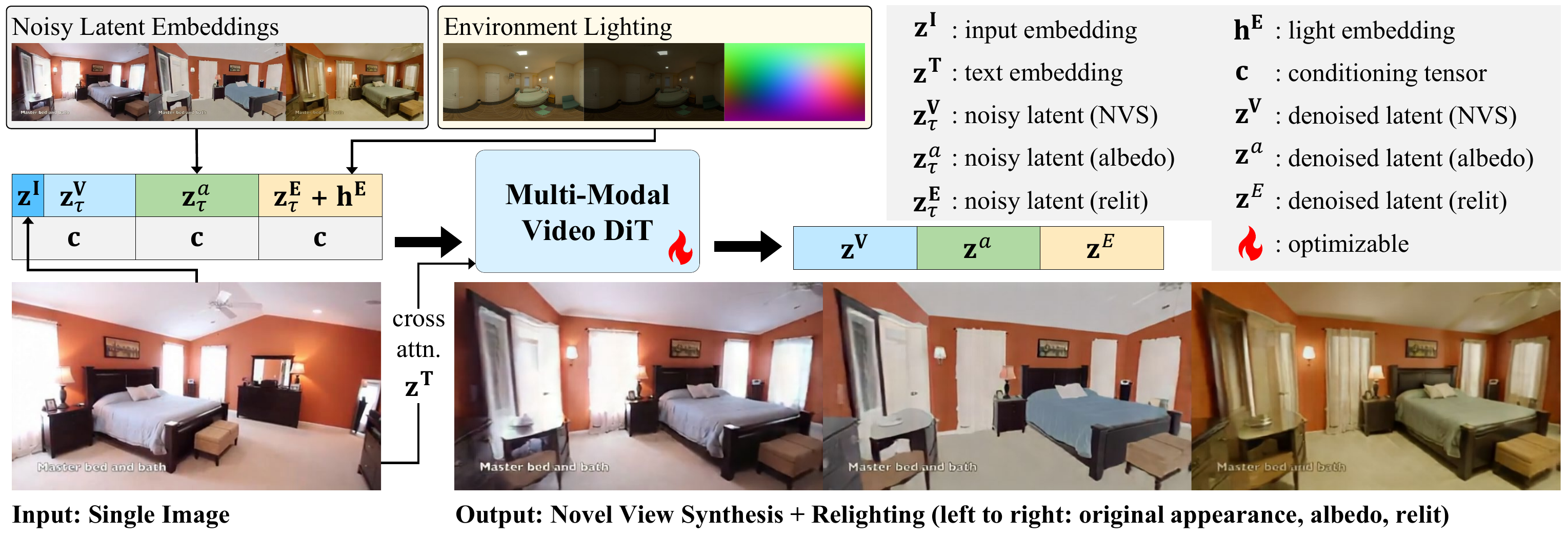}
    \caption{\textbf{An illustration of CamLit pipeline.} At the core of our framework is a multi-modal video DiT. This model takes as input a single RGB image, a camera trajectory, and an environment map. From these inputs, the model simultaneously generates a spatially and temporally aligned triplet of videos: (i) an RGB novel-view sequence under the same illumination as the input image, (ii) the corresponding relit sequence (with full shading from the environment map), and (iii) an albedo sequence capturing the scene’s intrinsics without shading. 
    }
    \label{fig:pipeline}
\end{figure*}

In this section, we detail our joint model design (\Cref{subsec:model}), data curation pipeline (\Cref{subsec:data}), and training and inference strategies for our model (\Cref{subsec:training}).

\subsection{Model Design}
\label{subsec:model}
\paragraph{Diffusion Backbone.}
Given an input RGB image $\mathbf{I} \in \mathbb{R}^{H \times W \times 3}$ with camera intrinsics (represented as focal length $\mathbf{f} \in \mathbb{R}^2$), a sequence of target camera poses $\mathbf{P} \in \mathbb{R}^{L \times 4 \times 4}$, and a target lighting environmental map $\mathbf{E} \in \mathbb{R}^{H \times W \times 3}$, CamLit aims to jointly generate three temporally-aligned video sequences: an NVS video $\mathbf{V}$ preserving the original lighting, a corresponding albedo video $\mathbf{a}$ capturing intrinsic material properties, and a relit video $\mathbf{V_E}$ rendered under the target lighting $\mathbf{E}$. Here, $H$, $W$ represents the spatial resolution and $L$ is the number of frames.
\Cref{fig:pipeline} illustrates an overview of our model design, which is based on a video diffusion model comprising a Video VAE $(\mathcal{E}, \mathcal{D})$ and a DiT $\mathcal{F}_{\theta}$~\cite{peebles2022dit} where $\theta$ denotes its trainable parameters.

Following the paradigm of Latent Diffusion Models~\cite{rombach2021highresolution}, the encoder $\mathcal{E}$ maps videos separately into three distinct latent embeddings $\{\mathbf{z}^{\mathbf{V}}, \mathbf{z}^{\mathbf{a}}, \mathbf{z}^{\mathbf{E}}\} \in \mathbb{R}^{l \times h \times w \times C}$, where $l=\frac{L-1}{8}+1$, $h=\frac{H}{8}$, and $w=\frac{W}{8}$ denote the spatial–temporal resolutions after $8\times$ downsampling by the encoder, and $C=16$ is the feature dimension.
These embeddings are subsequently perturbed with noise and processed by DiT to predict their denoised counterparts $\hat{\textbf{z}}^{\mathbf{V}}(\theta)$, $\hat{\textbf{z}}^{\mathbf{a}}(\theta)$, and $\hat{\textbf{z}}^{\mathbf{E}}(\theta)$ under the guidance of multi-modal conditions $\mathbf{c}$ and $\mathbf{z}^{\mathbf{T}}$. Formally, this denoising process can be written as:
\begin{equation}
\label{eq:diffusion}
    \hat{\textbf{z}}^{\mathbf{V}}(\theta), \hat{\textbf{z}}^{\mathbf{a}}(\theta), \hat{\textbf{z}}^{\mathbf{E}}(\theta) = \mathcal{F}_{\theta}(\mathbf{z}_\tau; \mathbf{c}, \mathbf{z}^{\mathbf{T}}, \tau)\,,
\end{equation}
where $\mathbf{z}_\tau$ denotes the fused noisy latent representation of the video triplet and associated environment maps at diffusion noise level $\tau$, and $\mathbf{c}$ and $\mathbf{z}^{\mathbf{T}}$ represent the multi-modal conditioning embeddings, which encode camera parameters and other guidance cues. In the following, we will delve into the details of the construction of these embeddings.

\paragraph{Latent Embedding with Environment Maps.}
To jointly predict the three target videos, we concatenate the noisy latent embeddings
${\mathbf{z}^{\mathbf{V}}_\tau, \mathbf{z}^{\mathbf{a}}_\tau, \mathbf{z}^{\mathbf{E}}_\tau}$
along the temporal dimension, forming three contiguous video chunks. To ensure spatial and temporal alignment across these chunks, we incorporate positional embeddings that combine rotary positional embeddings (RoPE)~\cite{su2024roformer} and learnable positional embeddings~\cite{cosmos}.
The same positional encoding is applied to all three video embeddings, so that corresponding tokens across modalities are aligned both spatially and temporally.
This design allows the joint prediction process to naturally resemble a video-to-video translation task, encouraging mutual learning and consistency among the multi-representation outputs.

For the environment maps, we follow prior work~\cite{DiffusionRenderer, he2025unirelight} and transform them into a low dynamic range (LDR) space compatible with the VAE pretraining domain.
Specifically, each environment map $\mathbf{E}$ is preprocessed into three buffers: an LDR map $\mathbf{E}_{\text{ldr}}$ obtained via Reinhard tone mapping~\cite{reinhard}; a normalized log-intensity map $\mathbf{E}_{\text{log}} = \log(\mathbf{E}+1)/E_{\text{max}}$ following~\cite{jin2024neural_gaffer}; and a direction map $\mathbf{E}_{\text{dir}}$ in which each pixel encodes a unit vector representing the corresponding light direction.
All buffers are resized to match the input video resolution. These processed buffers are then encoded by the VAE encoder $\mathcal{E}$ to yield the latent light embedding
\begin{equation}
    \mathbf{h}^{\mathbf{E}}=[\mathcal{E}(\mathbf{E}_{\text{ldr}});\mathcal{E}(\mathbf{E}_{\text{log}});\mathcal{E}(\mathbf{E}_{\text{dir}})] \in \mathbb{R}^{l \times h \times w \times 3C}\,,
\end{equation}
where $[\cdot]$ denotes the concatenation operation along the feature dimension. We concatenate $\mathbf{h}^{\mathbf{E}}$ with the latent embedding $\mathbf{z}^{\mathbf{E}}_\tau$ along the feature dimension, and pad the other latent embeddings $\mathbf{z}^{\mathbf{V}}_\tau$ and $\mathbf{z}^{\mathbf{a}}_\tau$ with zero-valued channels to maintain a consistent feature dimensionality across modalities.

Formally, the fused latent embedding $\mathbf{z}_\tau$ used in~\Cref{eq:diffusion} can be expressed as:
\begin{equation}
    \mathbf{z}_\tau = \text{Concat}([\mathbf{z}^{\mathbf{V}}_\tau; \mathbf{0}], [\mathbf{z}^{\mathbf{a}}_\tau; \mathbf{0}], [\mathbf{z}^{\mathbf{E}}_\tau; \mathbf{h}^{\mathbf{E}}]) + \mathbf{p}_{\text{rope}} + \mathbf{p}_{\text{learn}}\,,
\end{equation}
where $\text{Concat}(\cdot)$ denotes the concatenation along the temporal dimension.

\paragraph{Multimodal Conditioning with Camera Poses.}
To condition the denoising process on camera trajectory, we follow prior work~\cite{he2024cameractrl, zhou2025stable} and represent camera poses using Pl\"{u}cker embeddings~\cite{plucker1865xvii}: $\mathbf{r} = (\mathbf{d}, \mathbf{m}) \in \mathbb{R}^{L \times h \times w \times 6}$, where $\mathbf{d}$ denotes the per-pixel ray direction, and $\mathbf{m}$ is the moment vector obtained by taking the cross product of each ray direction and camera position. Considering the temporal compression ratio of VAE, we group and concatenate consecutive $8$ frames of Pl\"{u}cker embeddings, reshaping the tensor $\mathbf{r}$ into $l \times h \times w \times 48$.

We further employ binary condition masks $\mathbf{M}_{\text{cond}} \in \mathbb{R}^{l \times h \times w}$ to indicate which latent corresponds to the input image condition, and one-hot modality masks $\mathbf{M}_{\text{mod}} \in \mathbb{R}^{l \times h \times w \times 3}$ to distinguish the three target modalities: novel view synthesis, albedo prediction, and relighting.
Both masks are concatenated with the Pl\"{u}cker embeddings along the feature dimension, forming the complete conditioning embedding $\mathbf{c}$:
\begin{equation}
    \mathbf{c} = [\mathbf{r}; \mathbf{M}_{\text{cond}}; \mathbf{M}_{\text{mod}}] \in \mathbb{R}^{l \times h \times w \times 52}.
\end{equation}
This conditioning embedding $\mathbf{c}$ is then concatenated to each modality's latent embedding along the feature dimension, as shown in~\Cref{fig:pipeline}.

\paragraph{Context-Guided Diffusion.}
Beyond environment embeddings and multimodal conditioning with camera, we also incorporate textual context to improve the fidelity and semantic consistency of generated results.
Following the camera-conditioned foundation model described in~\cite{nvidia2025cosmosworldfoundationmodel}, we extract textual descriptions $\mathbf{T}$ from the input image using Pixtral-12B~\cite{agrawal2024pixtral12b}.
The extracted text is then encoded into a latent embedding $\mathbf{z}^{\mathbf{T}}$ using Google’s T5 Tokenizer~\cite{2020t5}.
This text embedding is injected into DiT via cross-attention, providing high-level semantic guidance without introducing external information.

\subsection{Data Curation}
\label{subsec:data}
\textbf{Video Triplets.}
Training our model requires a large-scale dataset of spatially and temporally aligned video triplets, comprising the original video, its intrinsic albedo reconstruction, and a corresponding relit version. However, such triplets are not readily available in existing datasets. To overcome this limitation, we construct a new dataset tailored for our task, enabling joint learning of video generation, albedo estimation, and relighting under consistent scene geometry and motion.

We start from RealEstate10K~\cite{zhou2018stereo}, a large-scale collection of in-the-wild videos covering diverse indoor and outdoor scenes. From this dataset, we curate $56,975$ video clips. 
To obtain the corresponding albedo and relit modalities, we leverage DiffusionRenderer~\cite{DiffusionRenderer}, a state-of-the-art video relighting framework capable of producing photometrically consistent inverse and forward renderings.
Specifically, we first employ the Inverse Renderer of DiffusionRenderer to generate per-frame G-buffer videos containing albedo, normals, depth, roughness, and metalness. Then, the Forward Renderer takes these G-buffers along with a randomly sampled HDR environment map from PolyHaven~\cite{polyhaven} to synthesize corresponding relit videos under novel illumination.

This process produces a total of $56,975$ triplets of original, albedo, and relit videos, each aligned in both space and time.
It is important to note that DiffusionRenderer is used only for data curation, not as part of our inference pipeline. Once trained, our model can operate from a single input image and does not rely on any external rendering models or video supervision at test time.

\paragraph{Camera Pose Normalization.}
Since our training videos are sourced from RealEstate10K~\cite{zhou2018stereo} with estimated camera poses of different scenes, it is essential to normalize these poses for numerical stability and to ensure that the model learns camera motion in a canonical coordinate space. Following prior work~\cite{jin2024lvsm, watson2024controlling, zhou2025stable, yu2024viewcrafter}, we perform a two-step normalization procedure.

First, for each video sequence, we compute the mean camera center across all frames and translate the camera poses so that this average center lies at the origin. We then rescale all camera positions such that their distances from the origin fall within a unit radius, normalizing the global scale of camera trajectories.
Second, to remove absolute pose bias and encourage learning of relative motion, we reparameterize all camera poses with respect to the first frame by making the first-frame camera pose an identity transformation.

\subsection{Training and Inference}
\label{subsec:training}
\paragraph{Training.} 
During training, the first frame of each video $\mathbf{V}$ is used as the input image $\mathbf{I}$. 
We denote the clean latent embeddings of the three modalities as 
$\mathbf{z}_{0}^{\mathbf{V}}$, $\mathbf{z}_{0}^{\mathbf{a}}$, and $\mathbf{z}_{0}^{\mathbf{E}}$, 
which serve as ground-truth targets for the denoising model. 
At each optimization step, Gaussian noise are added to the clean latent embeddings to obtain the corresponding noisy latents. 
To condition the model, the first noisy token of $\mathbf{z}_{\tau}^{\mathbf{V}}$ is replaced with the latent embedding $\mathbf{z}^{\mathbf{I}}$ of the input image $\mathbf{I}$. 
Following~\cite{cosmos}, a small amount of noise is added to $\mathbf{z}^{\mathbf{I}}$ to enhance robustness during inference. 
The denoised latent embeddings are then predicted using the diffusion process described in~\Cref{eq:diffusion}, and the trainable parameters of the DiT $\mathcal{F}_{\theta}$ are optimized by minimizing the following objective:
\begin{equation}
\begin{split}
\mathcal{L} = 
\Big\|\hat{\textbf{z}}^{\mathbf{V}}(\theta) - \textbf{z}_{0}^{\mathbf{V}} \Big\|_2^2 + \Big\|\hat{\textbf{z}}^{\mathbf{a}}(\theta) - \textbf{z}_{0}^{\mathbf{a}} \Big\|_2^2 + \Big\|\hat{\textbf{z}}^{\mathbf{E}}(\theta) - \textbf{z}_{0}^{\mathbf{E}} \Big\|_2^2\,.
\end{split}
\label{eq:objective}
\end{equation}
%
Additional training configurations and details can be found in the Cosmos technical report~\cite{cosmos}.

\paragraph{Inference.} 
At inference time, we iteratively apply $\mathcal{F}_{\theta}$ to Gaussian noise samples to generate a triplet of latent embeddings 
$\{\mathbf{z}^{\mathbf{V}}, \mathbf{z}^{\mathbf{a}}, \mathbf{z}^{\mathbf{E}}\}$, which are finally decoded back to videos through the decoder $\mathcal{D}$. 
We adopt classifier-free guidance (CFG)~\cite{ho2022classifier} using the text embedding $\mathbf{z}^{\mathbf{T}}$ as the conditional input. 
For the unconditional branch of CFG, instead of zero-padding the context embedding, we employ a negative prompt~\cite{ban2024understanding} with a fixed text description $\mathbf{T}$ that depicts a scene of poor quality, 
which helps suppress undesired artifacts and enhances generation fidelity.
\section{Experiments}
\label{sec:experiments}

\subsection{Implementation Details}
We fine-tune the full set of parameters of the \textbf{Cosmos-Predict1-7B-Video2World} foundation model~\cite{cosmos}, a DiT-based video diffusion model pretrained for future-frame prediction.

For training data, we use the RealEstate10K dataset~\cite{zhou2018stereo} preprocessed by PixelSplat~\cite{charatan2024pixelsplat}, which primarily contains videos at a resolution of $360\times640$.
All frames are resized to $352\times640$ to ensure divisibility by $32$ for feature encoding, and each training sample consists of $49$ consecutive frames per modality (original, albedo, and relit videos) for the balance between temporal context and computational efficiency.
Thus, each input clip has shape $(L, H, W) = (49, 352, 640)$.

We employ \texttt{Cosmos-Tokenize1-CV8x8x8-720p} as the video tokenizer to encode video frames into the latent space of DiT.
For each clip, the tokenizer maps the first frame to the first latent token and compresses every subsequent $8$ frames into the next temporal token, with an additional spatial compression factor of $8$.
Consequently, the encoded latent embeddings have dimensions $(l, h, w, C) = (7, 44, 80, 16)$.

Training is performed with a batch size of $32$ using the AdamW optimizer~\cite{loshchilov2017decoupled}, with a learning rate of $5\times10^{-5}$ and a weight decay of $0.1$.
The model is totally trained for $24{,}000$ iterations with BF16 mixed precision, taking approximately $3$ days on $32$ NVIDIA A100 GPUs.

During inference, we adopt $35$ denoising steps with the EDM scheduler~\cite{Karras2022edm}, and apply CFG with a guidance scale of $7$ for text prompts.

\subsection{Results}

\begin{figure*}[t]
    \centering
    \includegraphics[width=\textwidth]{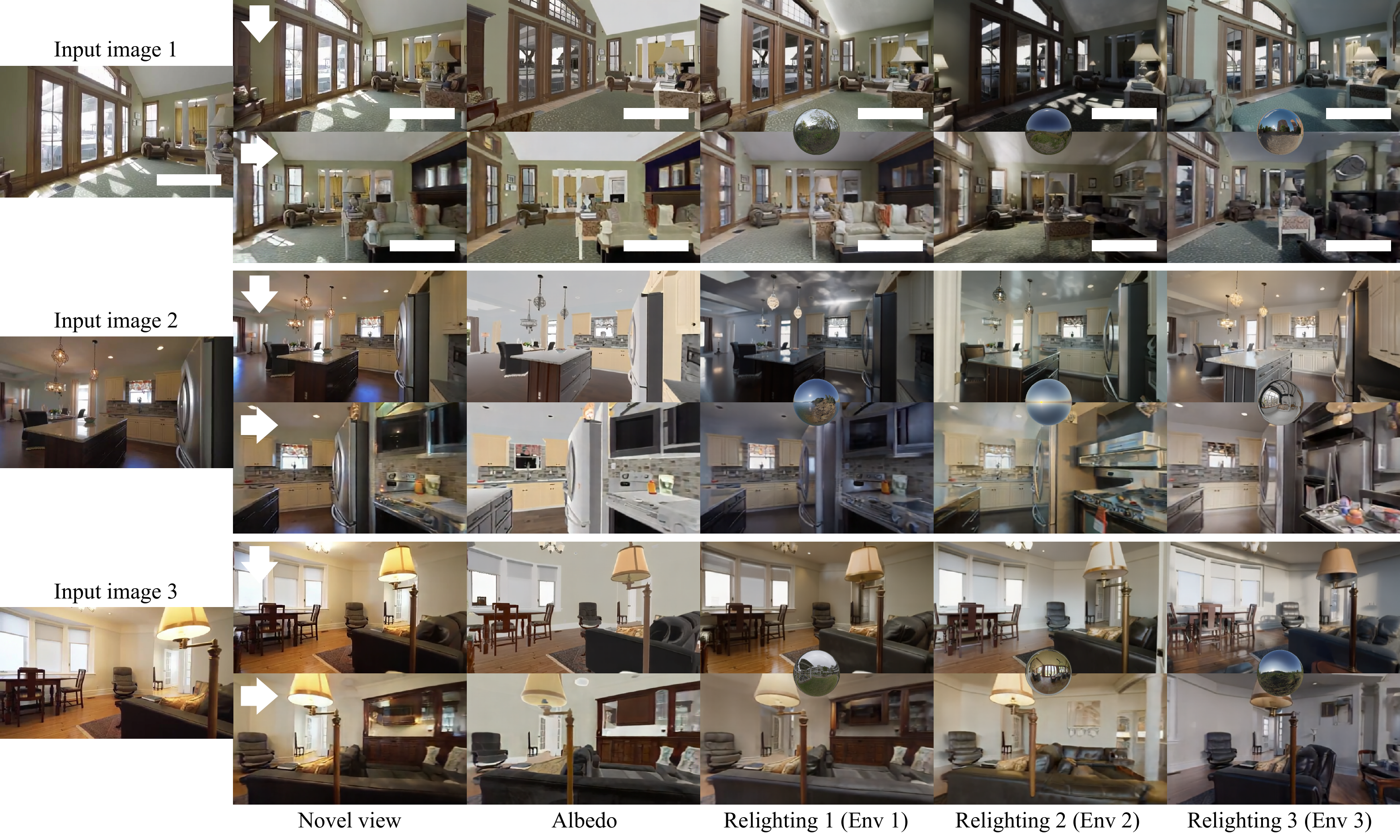}
    \caption{
    \textbf{Video generation examples of CamLit.} For each example, we visualize two camera trajectories, moving backward and turning right as indicated by the arrows, to reveal generated content in unseen regions. From left to right, we show the input image, a novel view frame under the original lighting, the corresponding albedo, and three relit novel view frames. The environment maps used for relighting are shown in the insets.
    }
    \label{fig:main_results}
\end{figure*}

As shown in~\Cref{fig:main_results}, CamLit generates high-quality novel-view and relit videos from a single input image. We show two camera trajectories, moving backward and turning right, to demonstrate CamLit's ability to predict plausible content in previously unseen regions. By conditioning on the target camera trajectory and different environment maps, CamLit produces diverse, photorealistic sequences for both indoor and outdoor scenes, making it practically useful for scalable video data generation.

\begin{table}[t]
    \centering
    \caption{FID scores computed on RealEstate10K~\cite{zhou2018stereo} test split. Our unified architecture performs NVS and relighting jointly, yet achieves fidelity on par with state-of-the-art approaches dedicated to each task.
    } 
    \label{tab:fid}
    \begin{tabular}{lcc}
        \toprule
        NVS Methods & FID $\downarrow$ \\
        \midrule
        SEVA~\cite{zhou2025stable} & 5.615 \\
        GEN3C~\cite{ren2025gen3c} & 5.144 \\
        Our nvs-only & 6.724 \\
        Ours & 7.213 \\
        \midrule
        Relighting Methods & FID $\downarrow$ \\
        \midrule
        DiffusionRenderer~\cite{DiffusionRenderer} & 14.088 \\
        Ours & 13.731 \\
        \bottomrule
    \end{tabular}
\end{table}

\begin{table}[t]
    \centering
    \caption{FVD scores computed on RealEstate10K~\cite{zhou2018stereo} test split. Our unified system achieves video quality comparable to state-of-the-art approaches dedicated to each task.
    } 
    \label{tab:fvd}
    \begin{tabular}{lcc}
        \toprule
        NVS Methods & FVD $\downarrow$ \\
        \midrule
        SEVA~\cite{zhou2025stable} & 94.95 \\
        GEN3C~\cite{ren2025gen3c} & 91.69 \\
        Our nvs-only & 78.03 \\
        Ours & 99.17 \\
        \midrule
        Relighting Methods & FVD $\downarrow$ \\
        \midrule
        DiffusionRenderer~\cite{DiffusionRenderer} & 103.98 \\
        Ours & 109.27 \\
        \bottomrule
    \end{tabular}
\end{table}

\subsection{Comparisons}

To the best of our knowledge, no existing video generation method provides simultaneous control over both camera motion and scene illumination.
Given our objective is to develop a more versatile model that unifies these two conditioning dimensions, prior approaches that are typically designed for either camera- or lighting-conditioned generation cannot cover the full spectrum of our method.
Consequently, we evaluate our model on each task independently, against state-of-the-art baselines in their respective domains.
For NVS, we compare with SEVA~\cite{zhou2025stable} and GEN3C~\cite{ren2025gen3c}, two high-quality, camera-conditioned video diffusion models.
For relighting, we benchmark against DiffusionRenderer (DR)~\cite{DiffusionRenderer}, one of the strongest publicly available video diffusion models for relighting.

For quantitative evaluation, we focus on image-to-video generation from a single input frame, rather than exact scene reconstruction, since faithfully recovering unseen regions is inherently ill-posed.
We therefore adopt the Fréchet Inception Distance (FID)~\cite{heusel2017gans} and Fréchet Video Distance (FVD)~\cite{unterthiner2019fvd} computed on the RealEstate10K test split as the primary metrics to assess perceptual generation quality and temporal consistency.

\paragraph{Novel View Synthesis.}
For NVS evaluation, we design four canonical camera trajectories: \emph{move forward}, \emph{move backward}, \emph{turn left}, and \emph{turn right}. We set the translation distance to $2$ units for the first $2$ trajectories, and the rotation angle to $45$ degrees for the last $2$ ones, yielding natural camera motion over $49$ frames. 
We randomly select $1,280$ scenes from the RealEstate10K test split, using the first frame of each scene as the input image.
For each scene, we pair the input with a randomly chosen HDR environment map from PolyHaven~\cite{polyhaven} and generate four triplets of NVS, albedo, and relit videos conditioned on the respective camera trajectories.
This setup yields a total of $5,120$ triplets, each containing a $49$-frame video sequence.

For SEVA and GEN3C, we perform inference on the same $1,280$ input images using the identical four trajectories to ensure fair comparison.
Since their default outputs contain $112$ and $121$ frames, respectively, we uniformly sample $49$ frames from their turning-camera results for side-by-side evaluation.
For moving-camera trajectories, single-view NVS models inherently suffer from scale ambiguity, which can cause different apparent motion magnitudes for the same translation.
Empirically, we find that the first $49$ frames of most SEVA and GEN3C outputs exhibit motion comparable to our generated results, and we therefore use these frames for evaluation.

\begin{figure}[t]
    \centering
    \includegraphics[width=\columnwidth]{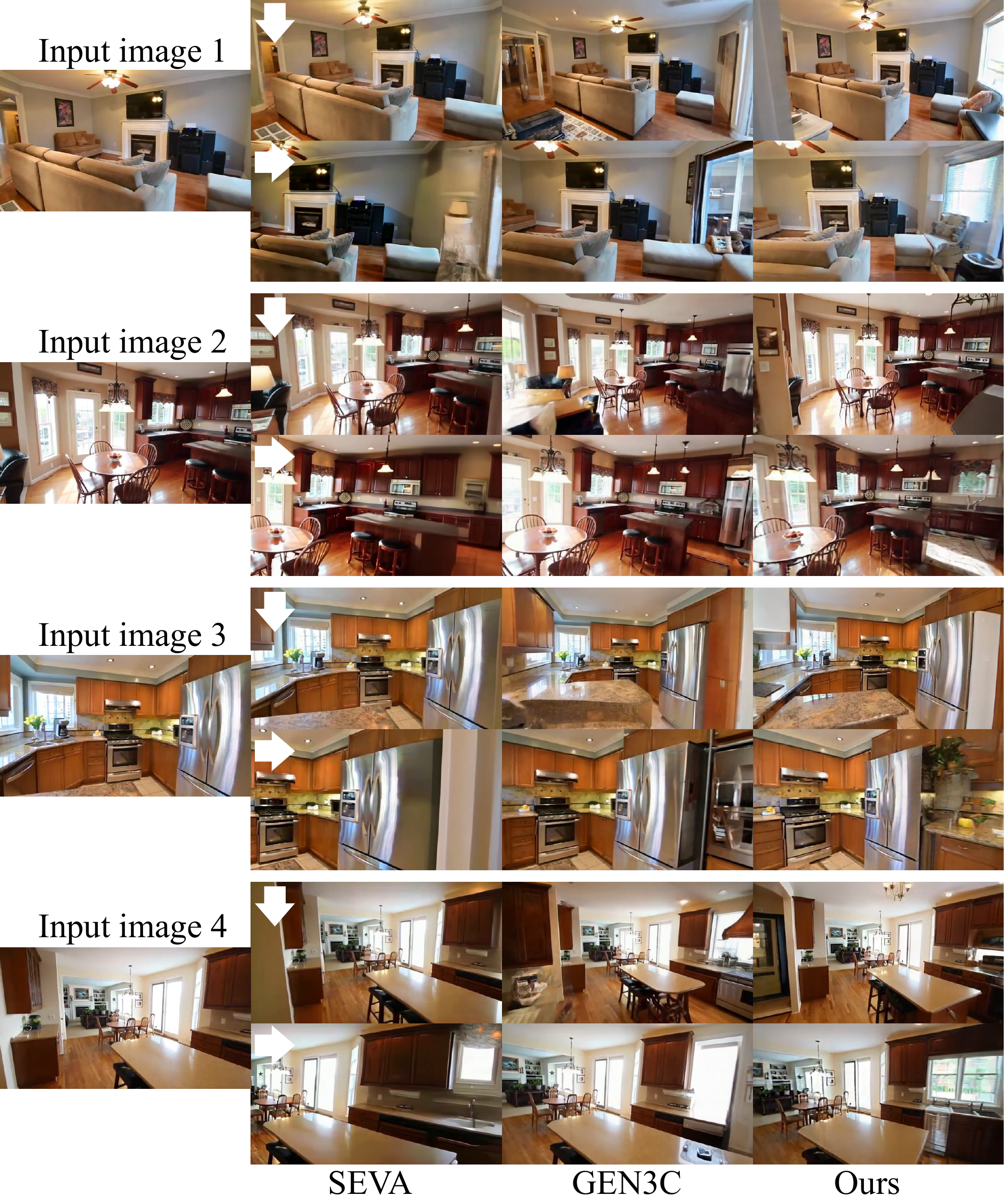}
    \caption{
    \textbf{Qualitative comparison of novel view synthesis methods.} For each input image, we apply two camera trajectories, moving backward ($1$st row) and turning right ($2$nd row), as indicated by the arrows. Our model, which performs NVS and relighting jointly, achieves NVS quality on par with state-of-the-art methods specifically dedicated to NVS.
    }
    \label{fig:nvs_cmp}
\end{figure}

As illustrated in~\Cref{fig:nvs_cmp}, our model synthesizes high-quality content in previously unseen regions, which blends seamlessly with the visible areas of the input image while maintaining consistent geometry and appearance.
Under camera motion, SEVA exhibits noticeable geometric jittering, whereas GEN3C and our method produce more stable and detailed results.
Please refer to our supplementary video for full visual comparisons.
Quantitative results are summarized in~\Cref{tab:fid,tab:fvd}, where all three methods (SEVA, GEN3C, and ours) achieve comparable FID and FVD values, confirming that our unified camera- and lighting-conditioned generation framework preserves the fidelity of NVS outputs.

\begin{figure}[t]
    \centering
    \includegraphics[width=\columnwidth]{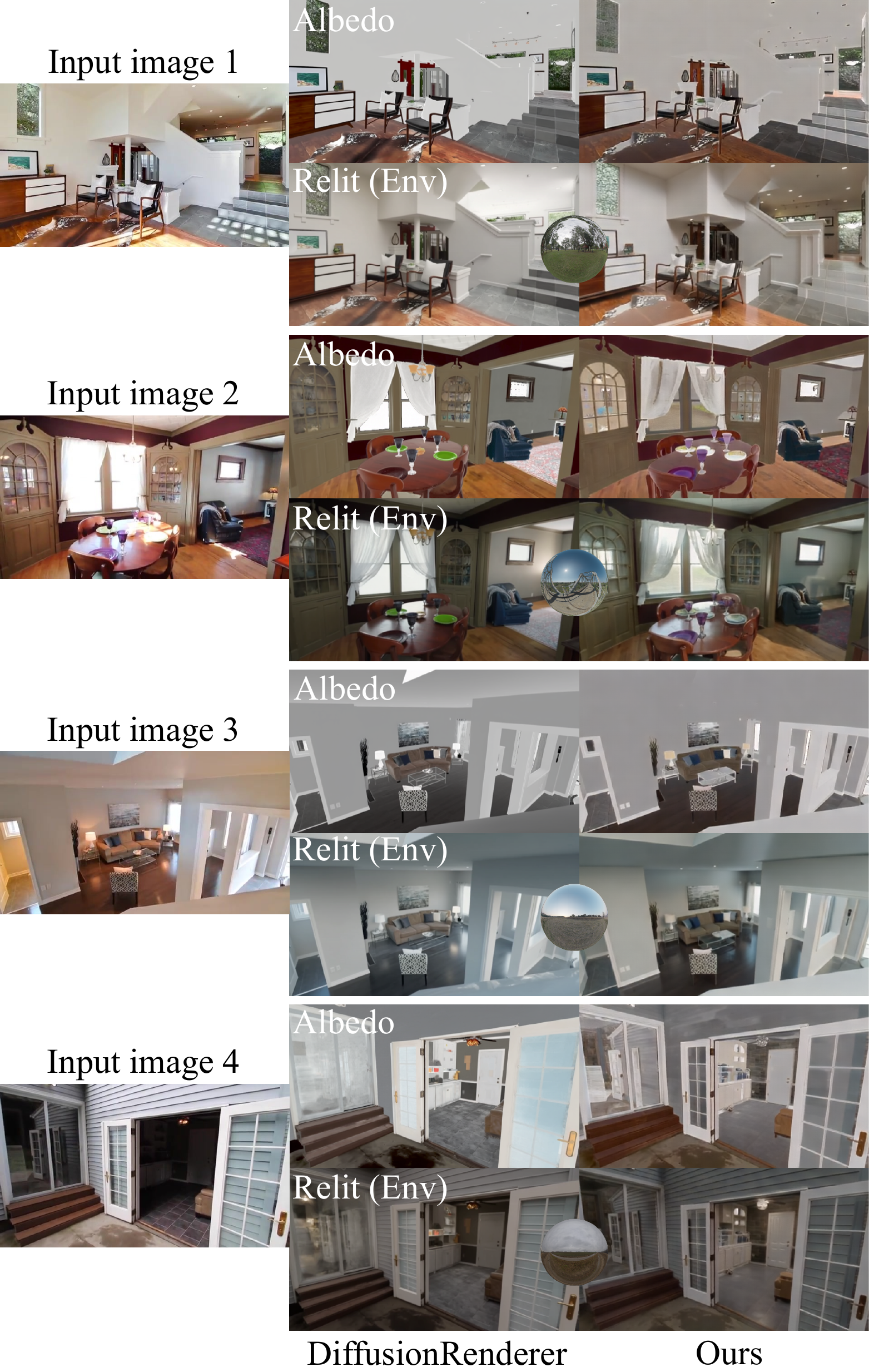}
    \caption{
    \textbf{Qualitative comparison of relighting methods.} Our approach produces albedo and relit videos with quality comparable to DiffusionRenderer~\cite{DiffusionRenderer}, which is our theoretical performance upper bound. The environment maps used for relighting are shown in the middle insets.
    }
    \label{fig:relit_cmp}
\end{figure}

\paragraph{Relighting.} 
For the relighting evaluation, we use the same $1{,}280$ scenes as in the NVS experiments.
For each scene, we feed the ground-truth video frames into DR to generate corresponding albedo and relit videos, using the same environment maps employed in our inference.
As illustrated in~\Cref{fig:relit_cmp}, our method produces albedo and relit results that are visually comparable to those of DR.
The generated albedo videos from both approaches effectively remove the illumination from the input frames and exhibit consistent, plausible colors across semantic regions.
Likewise, both relit videos preserve the intrinsic scene properties -- geometry, materials, and spatial layout -- while generating realistic rendering under the designated environment maps.
As reported in~\Cref{tab:fid,tab:fvd}, we obtain FID and FVD scores similar to DR, again demonstrating that we preserve the fidelity of relighting without compromising generalization.

\subsection{Ablation Study}
\begin{figure}[t]
    \centering
    \includegraphics[width=\columnwidth]{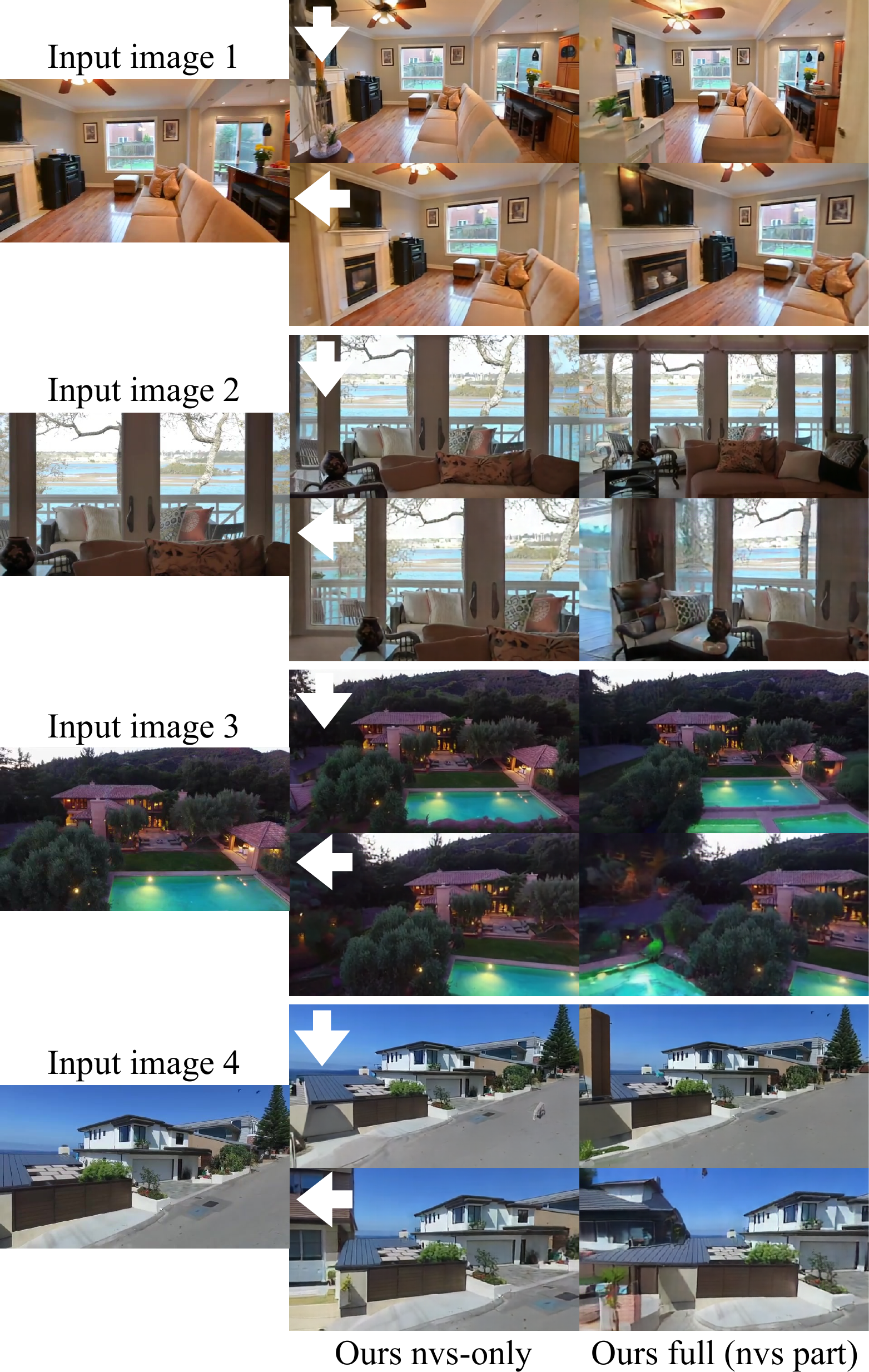}
    \caption{
    \textbf{Qualitative comparison of our NVS‑only and full models.} Two camera trajectories, moving backward and turning left as indicated by arrow directions. Both models produce high‑quality, geometrically consistent novel views, demonstrating that the unified formulation preserves NVS performance while integrating relighting.
    }
    \vspace{-0.7em}
    \label{fig:ablation_nvsonly}
\end{figure}

Our model unifies NVS and relighting within a single architecture, achieving joint control over camera and illumination without degrading performance on either task.
As verified by the relighting comparison in~\Cref{fig:relit_cmp},~\Cref{tab:fid,tab:fvd}, incorporating NVS into the framework does not compromise relighting quality.
To further assess the impact on NVS, we train an NVS-only baseline using the same NVS training data and configuration as our full model.
This baseline takes the first frame and target camera trajectory as inputs and predicts future NVS frames, but omits albedo and relighting branches.
As shown in~\Cref{fig:ablation_nvsonly}, ~\Cref{tab:fid,tab:fvd}, the NVS results from our unified model are nearly indistinguishable from those of the NVS-only baseline -- both qualitatively and in terms of FID and FVD.
Both models generate high-quality, geometrically consistent novel views, demonstrating that our unified formulation preserves NVS performance while successfully integrating relighting capability.

\section{Limitations}
\label{sec:limitations}

\paragraph{Entanglement of NVS and Relighting.}
Because our approach adopts a multimodal joint denoising formulation, NVS and relighting remain intrinsically coupled in our formulation.
Consequently, it is not guaranteed to generate perfectly identical novel-view sequences while varying the illumination only.
Although this coupling is acceptable for many practical scenarios such as controllable video generation or data augmentation, it limits the applicability of our method in settings that require fully disentangled control over lighting and geometry.

\paragraph{Lack of Explicit Light-Source Control.}
We represent illumination using environment maps, which efficiently encode global lighting and facilitate conditioning across scenes.
However, this representation does not allow explicit manipulation of individual light emitters.
For example, toggling a specific lamp on or off cannot be deterministically achieved, which may only emerge implicitly as the environment map changes in our current method.
Integrating explicit light-source modeling and control~\cite{magar2025lightlab} with joint novel view synthesis remains an interesting direction for future research.

\paragraph{Sensitivity to Extreme Camera Motions.}
CamLit can produce suboptimal results under camera trajectories that are underrepresented in the training set, such as large yaw rotations of around $90$ degrees.
We expect this limitation to diminish with training on datasets containing more diverse and aggressive camera motions, \eg, DL3DV-10K~\cite{ling2024dl3dv}.

\section{Conclusion}
\label{sec:conclusion}

We presented CamLit, a unified video diffusion model for simultaneous novel view synthesis and relighting from a single image. By conditioning on user-specified camera trajectories and an environment map, CamLit generates spatially and temporally consistent videos with explicit control over viewpoint and illumination. Within a single denoising process, our model produces photorealistic novel views alongside corresponding intrinsic albedo and relit frames, ensuring cross-modal consistency in scene content and lighting. Training on a large set of synthetic video triplets derived from RealEstate10K~\cite{zhou2018stereo} using DiffusionRenderer~\cite{DiffusionRenderer} enabled the model to learn realistic lighting effects and handle diverse scenes without requiring multi-view input. Experimental results confirm that CamLit achieves high-fidelity outputs on par with state-of-the-art methods in both novel view synthesis and relighting, without sacrificing visual quality in either task. These findings demonstrate that a single generative model can effectively integrate camera and lighting control, simplifying the video generation pipeline while maintaining competitive performance and consistent realism.

\newpage

{
    \small
    \bibliographystyle{ieeenat_fullname}
    \bibliography{main}
}


\end{document}